\documentclass[a4paper,fleqn]{cas-dc}

\usepackage[numbers]{natbib}

\def\tsc#1{\csdef{#1}{\textsc{\lowercase{#1}}\xspace}}
\tsc{WGM}
\tsc{QE}

\begin{document}
\let\WriteBookmarks\relax
\def\floatpagepagefraction{1}
\def\textpagefraction{.001}

\shorttitle{ST-GNN - Predicting Cognitive Profiles from Structural MRI using Spatial Graph Representation Learning}    

\shortauthors{Shuva et al.}  

\title [mode = title]{\textbf{Predicting Cognitive Profiles from Structural MRI using Spatial Graph Representation Learning}}  

\author[1,2]{Taslima Ferdaus Shuva}[orcid=0000-0002-4672-4528]
\credit{Conceptualization, Methodology, Formal analysis, Investigation, Data Curation, Writing - Original Draft}

\author[3,4]{Md. Tanvir Rahman} [orcid=0000-0002-6173-0483]
\cormark[1]
\ead{tanvirrahman@mbstu.ac.bd}
\credit{Resources, Visualization, Project administration}

\author[1]{Risala Tasin Khan} [orcid=0000-0001-8236-5959]
\credit{Validation, Supervision, Writing - Review \& Editing}

\author[3,5,6]{Mohammad Ali Moni} [orcid=0000-0003-0756-1006]
\credit{Writing - Review \& Editing}

\author[7]{ASM Ashraf Mahmud}[orcid=0000-0002-3832-1163]
\cormark[1]
\ead{ashraf.mahmud@uws.ac.uk}
\credit{Writing - Review \& Editing, Funding acquisition}

\affiliation[1]{organization={Institute of Information Technology, Jahangirnagar University},
            addressline={Savar}, 
            city={Dhaka},
            postcode={1342}, 
            country={Bangladesh}}

\affiliation[2]{organization={Department of Computer Science and Engineering, Daffodil International University},
            addressline={Daffodil Smart City (DSC), Birulia, Savar}, 
            city={Dhaka},
            postcode={1216}, 
            country={Bangladesh}}

\affiliation[3]{organization={School of Health and Rehabilitation Sciences, The University of Queensland},
            state={QLD},
            postcode={4072}, 
            country={Australia}}

\affiliation[4]{organization={Department of Information and Communication Technology, Mawlana Bhashani Science and Technology University},
            city={Tangail},
            postcode={1902}, 
            country={Bangladesh}}

\affiliation[5]{organization={AI and Cyber Futures Institute, Charles Sturt University},
            city={Bathurst},
            postcode={2795}, 
            state={New South Wales},
            country={Australia}}

\affiliation[6]{organization={Rural Health Research Institute, Charles Sturt University},
            city={Orange},
            postcode={2800}, 
            state={New South Wales},
            country={Australia}}

\affiliation[7]{organization={Faculty of Computing Science and Engineering, University of the West of Scotland},
            addressline={High Street}, 
            city={Paisley},
            postcode={PA1 2BE}, 
            country={United Kingdom}}


\begin{abstract}
Predicting individual differences in cognitive abilities from structural neuroimaging remains challenging due to the high dimensionality and geometric complexity of brain morphology. Conventional machine learning typically flattens regional morphometry into unstructured tabular features, thereby failing to explicitly represent spatial relationships among brain regions. In this study, we propose a Spatial-Topological Graph Neural Network (ST-GNN) to predict four distinct cognitive phenotypes (cognitive flexibility, inhibitory control, working memory, and processing speed) across target-specific cohorts ($N = 2,378 - 2,424$). By projecting FreeSurfer-derived regional morphometry onto a population-level standardized spatial graph, this approach restricts message passing to spatially defined neighboring regions. Systematic evaluations demonstrate that the ST-GNN effectively fuses spatial morphology with systemic covariates, outperforming conventional linear baselines and topology-free deep learning models. Critically, extensive ablations reveal that imposing a sparse, localized geometric topology ($K$-nearest neighbors) and utilizing global feature scaling—which preserves absolute morphometric scale rather than flattening the anatomical hierarchy—outperform unconstrained fully connected networks and region-normalized approaches. These design choices provide highly transferable architectural inductive biases for spatial engineering tasks, while revealing differences between computational routing patterns and morphometric feature attribution. Integrated Gradients analysis identified distinct morphological features with relatively high attribution across the cognitive targets, with several findings being broadly consistent with prior neuroanatomical evidence, such as cerebellar volume for working memory and superior frontal surface area for general executive function. These findings suggest that standardized structural MRI may provide useful information for task-specific cognitive prediction and establish a foundation for future investigation of MRI-based neurodevelopmental assessment.
\end{abstract}


\maketitle

\begin{table*}[t]
\centering
\caption{Participant characteristics for the target-specific cohorts included in the family-aware cross-validation analysis}
\label{tab:demographics_imputed}
\renewcommand{\arraystretch}{1.15}
\begin{tabular*}{\textwidth}{@{\extracolsep{\fill}} l c c c c @{}}
\toprule
\multirow{2}{*}{\textbf{Characteristic}} & \multicolumn{4}{c}{\textbf{Cognitive Targets}} \\
\cmidrule{2-5}
& \textbf{Card Sort} & \textbf{Flanker} & \textbf{List Sort} & \textbf{Processing Speed} \\
\midrule
\multicolumn{5}{l}{\textbf{Sample Information}} \\
\hspace{3mm} Subjects ($n$) & 2,424 & 2,422 & 2,378 & 2,398 \\
\hspace{3mm} Families ($n$) & 2,424 & 2,422 & 2,378 & 2,398 \\
\addlinespace
\multicolumn{5}{l}{\textbf{Demographics \& Anthropometrics}} \\
\hspace{3mm} Age (years) & 10.5 $\pm$ 3.5 & 10.5 $\pm$ 3.5 & 10.6 $\pm$ 3.4 & 10.5 $\pm$ 3.5 \\
\hspace{3mm} eTIV ($mm^3$) & 1,475,563 $\pm$ 174,136 & 1,475,539 $\pm$ 174,160 & 1,475,831 $\pm$ 174,475 & 1,475,323 $\pm$ 174,391 \\
\hspace{3mm} Sex \\
\hspace{6mm} Male & 1,532 (63\%) & 1,531 (63\%) & 1,502 (63\%) & 1,520 (63\%) \\
\hspace{6mm} Female & 892 (37\%) & 891 (37\%) & 876 (37\%) & 878 (37\%) \\
\hspace{3mm} Handedness \\
\hspace{6mm} Right & 2,137 (88\%) & 2,134 (88\%) & 2,097 (88\%) & 2,114 (88\%) \\
\hspace{6mm} Left & 287 (12\%) & 288 (12\%) & 281 (12\%) & 284 (12\%) \\
\addlinespace
\multicolumn{5}{l}{\textbf{Socioeconomic Status}} \\
\hspace{3mm} Parental Education Score & 17.9 $\pm$ 3.0 & 17.9 $\pm$ 3.0 & 17.9 $\pm$ 3.0 & 17.9 $\pm$ 3.0 \\
\hspace{3mm} Parental Occupation Score & 34.5 $\pm$ 9.2 & 34.5 $\pm$ 9.2 & 34.6 $\pm$ 9.1 & 34.5 $\pm$ 9.2 \\
\hspace{3mm} Financial Support \\
\hspace{6mm} Level 1.0 & 501 (21\%) & 501 (21\%) & 484 (20\%) & 494 (21\%) \\
\hspace{6mm} Level 2.0 & 265 (11\%) & 265 (11\%) & 259 (11\%) & 263 (11\%) \\
\hspace{6mm} Level 3.0 & 1,658 (68\%) & 1,656 (68\%) & 1,635 (69\%) & 1,641 (68\%) \\
\addlinespace
\multicolumn{5}{l}{\textbf{Acquisition Site}} \\
\hspace{3mm} Site 1 & 853 (35\%) & 854 (35\%) & 838 (35\%) & 848 (35\%) \\
\hspace{3mm} Site 2 & 4 (<1\%) & 4 (<1\%) & 4 (<1\%) & 4 (<1\%) \\
\hspace{3mm} Site 3 & 714 (29\%) & 714 (29\%) & 694 (29\%) & 708 (30\%) \\
\hspace{3mm} Site 4 & 852 (35\%) & 849 (35\%) & 841 (35\%) & 837 (35\%) \\
\hspace{3mm} Site 5 & 1 (<1\%) & 1 (<1\%) & 1 (<1\%) & 1 (<1\%) \\
\addlinespace
\multicolumn{5}{l}{\textbf{Target Scores}} \\
\hspace{3mm} Mean Target Score & 87.1 $\pm$ 19.5 & 86.1 $\pm$ 15.7 & 92.4 $\pm$ 17.8 & 88.6 $\pm$ 20.0 \\
\bottomrule
\end{tabular*}

\vspace{4pt}
\raggedright
\footnotesize \textit{Note:} Continuous variables are reported as mean $\pm$ standard deviation, while categorical variables are presented as frequencies and percentages. Target scores represent age-uncorrected metrics to preserve genuine developmental variability. To enforce a strict zero-leakage evaluation, missing data points for continuous (median) and categorical (mode) variables were explicitly imputed within each fold of the five-fold cross-validation scheme using only training-set parameters. Across all evaluation cohorts, missingness and subsequent test-fold imputation rates remained minimal: Socioeconomic status occupation ($\sim$6.1\%), financial support ($\sim$2.2\%), Socioeconomic status education ($\sim$1.3\%), and handedness (0.1\%). Subject and family counts are identical across all cognitive targets because the stringent multimodal quality control pipeline naturally eliminated at least one member of every identified sibling pair, resulting in evaluation cohorts composed exclusively of independent singletons.
\end{table*}

\section{Introduction}
Cognitive development is a multidimensional process supported by distributed and interacting brain systems. Rather than reflecting a single ability, cognition comprises distinct functions including cognitive flexibility, inhibitory control, attention, working memory, and processing speed \cite{Diamond2013}. Reliable measurement of these domains is therefore important for understanding individual differences in neurodevelopment and for establishing quantitative brain--behavior relationships. The NIH Toolbox Cognition Battery (NIHTB-CB) provides standardized measures of these complementary cognitive abilities across the developmental lifespan \cite{weintraub2013cognition,weintraub2013pediatric}. Specifically, the Dimensional Change Card Sort (DCCS) assesses cognitive flexibility, the Flanker Inhibitory Control and Attention Test measures inhibitory control and selective attention, List Sorting evaluates working memory, and Pattern Comparison measures processing speed \cite{zelazo2013executive,tulsky2014list,carlozzi2014pattern}. These four phenotypes provide a useful framework for investigating whether individual differences in brain structure can help predict distinct cognitive abilities.

From an engineering and computational modeling perspective, a central challenge is efficiently representing complex, high-dimensional topological data to predict behavioral variation without relying on computationally expensive or acquisition-sensitive procedures like diffusion tractography. Structural magnetic resonance imaging (sMRI), particularly T1-weighted MRI, provides non-invasive measures of cortical and subcortical morphology, including cortical thickness, surface area, and regional volume. Such measures have been associated with cognitive variation across development and adulthood \cite{bethlehem2022brain}, motivating predictive approaches that move beyond population-level correlations. Large neuroimaging cohorts have further enabled the use of machine-learning models to quantify individual differences in cognition. However, brain--behavior associations remain difficult to access because effect sizes are often small relative to between-subject variability, emphasizing the importance of large samples, rigorous validation, and careful control of confounding factors \cite{marek2022reproducible,scheinost2019rules}.

Previous neuroimaging studies have demonstrated that cognitive phenotypes can be predicted from brain organization, although much of this literature has emphasized functional connectivity, composite cognitive measures, or multimodal representations \cite{rahman2026understanding}. Studies using neuroimaging cohorts have predicted NIH Toolbox measures, including DCCS, Flanker, List Sorting, and Pattern Comparison, from functional connectivity and related connectomic representations \cite{he2020deep,genon2022linking,cui2020individual}. Structural neuroimaging has been associated with these cognitive domains within the Healthy Brain Network (HBN), including studies that connect regional anatomical differences to NIH Toolbox performance \cite{begue2024cerebellum}. More recently, explainable machine-learning analyses have demonstrated prediction of multiple NIH Toolbox cognitive assessments from diffusion MRI-derived white-matter features, further supporting the relationship between brain structure and individual cognitive performance \cite{lo2025shape}. More broadly, predictive studies using structural and functional connectivity have demonstrated reproducible brain-related variation in cognitive performance, but many rely on multimodal, connectivity-based, or composite representations rather than task-specific prediction from regional structural morphometry alone \cite{rahman2026understanding}. This raises an important question: to what extent can distinct cognitive abilities be predicted from regional structural morphometry using a representation that preserves the brain's spatial organization?

A major limitation of conventional sMRI-based prediction is that regional morphometric measures are commonly flattened into independent tabular features. Although effective for conventional statistical and machine-learning models, this representation does not explicitly encode relationships between anatomically neighboring regions. The problem is not simply dimensionality; brain morphology is spatially organized, and predictive signals may arise from distributed patterns across anatomically related regions. Large-scale comparisons of machine-learning approaches have shown that deep models do not necessarily provide the gains observed in conventional machine-learning datasets when applied to neuroimaging data \cite{schulz2020different}. Consequently, an important methodological question is not simply whether a more complex predictor can be applied to sMRI, but whether the input representation appropriately reflects the spatial organization of neuroanatomy.

Graph neural networks (GNNs) provide such a framework by representing brain regions as nodes and their relationships as edges, allowing localized message passing to jointly model regional features and their topological context \cite{bessadok2023graph,cui2023braingb}. Most graph-based neuroimaging approaches, however, construct edges from functional connectivity, diffusion-derived structural connectivity, or other subject-specific connectomic measurements. Such approaches are powerful because their graphs can reflect individualized relationships between brain regions, but they also introduce additional imaging requirements and preprocessing complexity. In particular, diffusion MRI-based tractography can be computationally demanding and sensitive to acquisition quality, fiber geometry, and tractography methodology. A complementary strategy is therefore to construct a common spatial graph based on the proximity of brain regions, enabling graph-based learning from T1-weighted structural morphometry without requiring individualized diffusion tractography.

The use of a spatial graph, however, does not eliminate the methodological challenges of individual-difference prediction. The HBN provides a substantial cohort for studying individual differences in neurodevelopment but also exhibits heterogeneity in age, sex, socioeconomic characteristics, acquisition site, and family structure \cite{alexander2017open}. Related participants can introduce familial leakage when individuals from the same family are assigned to different data partitions, while preprocessing or normalization performed before cross-validation can leak information from the test data \cite{scheinost2019rules,snoek2019confounds}. Recent systematic evidence further indicates that deep-learning neurocognition research remains concentrated on a small set of major datasets, frequently lacks interpretability, and often employs inconsistent evaluation strategies, highlighting the need for rigorous validation, interpretable modeling, and broader methodological standardization \cite{rahman2026understanding}.

Applying graph neural networks to neuroanatomy raises important methodological considerations related to architectural priors. We explore how dense the anatomical graph should be—comparing sparse, localized routing with fully-connected message passing—and how to scale morphological features to mirror the brain's macro-structural hierarchy accurately. Additionally, interpretability remains crucial when predictive models aim to offer neurobiological insights. Achieving high predictive accuracy alone does not reveal which structural features influence the results, and traditional feature-importance measures may be hard to interpret due to correlated neuroanatomical and demographic variables \cite{kohoutova2020interpreting}. Attribution techniques like Integrated Gradients provide a rigorous method for assessing how input features contribute to deep-network outputs \cite{sundararajan2017axiomatic}.

Despite these advances, an important methodological gap remains in determining whether simple spatial organization of regional structural morphometry can improve individual-level cognitive prediction without relying on subject-specific functional or diffusion-derived connectivity. In particular, it remains unclear whether the predictive benefit of graph-based modeling arises from preserving spatial relationships among morphometric regions or simply from the increased flexibility of nonlinear models. This motivates a controlled comparison of spatially constrained graph representations with flattened structural representations, alternative graph densities, and conventional linear baselines.

In this study, we introduce a Spatial-Topological Graph Neural Network (ST-GNN) for predicting four distinct cognitive phenotypes (cognitive flexibility, inhibitory control and attention, working memory, and processing speed) from FreeSurfer-derived regional T1-weighted sMRI morphometry in the pediatric HBN cohort. Instead of constructing subject-specific connectivity graphs, we define a shared population-level spatial graph by linking anatomical regions according to their proximity in standardized Montreal Neurological Institute (MNI) space \cite{Fonov2011}. Individual cortical and subcortical morphometric features are projected onto this common topology, allowing graph-based message passing to capture local spatial context without requiring diffusion MRI. Demographic and acquisition-related variables are incorporated through a separate late-fusion pathway, while family-aware stratified cross-validation is used to prevent familial leakage and preserve target distributions across partitions. Finally, an Integrated Gradients-based explainability framework is used to quantify regional structural contributions and compare them with systemic covariates. This framework therefore combines four complementary components for pediatric cognitive prediction: task-specific phenotype modeling, spatially informed structural representation, leakage-controlled evaluation, and interpretable analysis of model predictions.

\begin{figure*}[t]
    \centering
    \includegraphics[width=\textwidth]{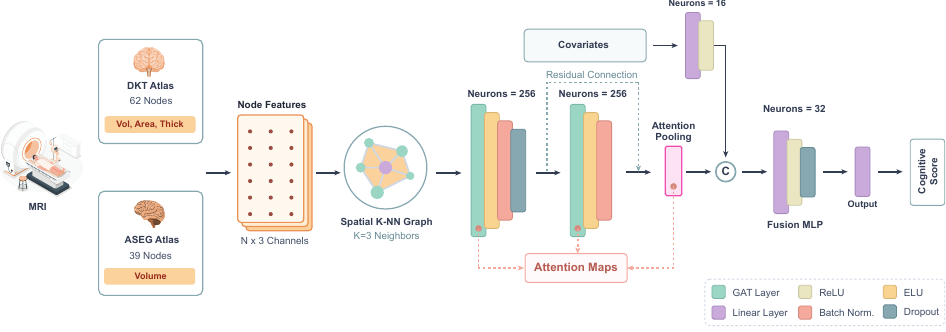}
    \caption{Architecture of the proposed Spatial-Topological Graph Neural Network (ST-GNN) for cognitive score prediction from structural MRI and systemic covariates. The left panel illustrates the spatial graph construction, in which the 101 anatomical regions defined by the DKT and ASEG atlases are embedded in a spatial K-nearest-neighbor graph with $K=3$. Morphological node features are combined with learnable node-identity embeddings to form the initial node representation. The middle panel depicts two graph-attention blocks with residual connections and regularization, followed by learned global attention pooling to obtain a whole-brain structural representation. The right panel shows the late-fusion stage, where demographic, socioeconomic, and acquisition-related covariates are processed through a parallel multilayer perceptron and combined with the graph representation to predict the continuous cognitive score.}
    \label{fig:gnn_methodology}
\end{figure*}

\begin{figure*}[t]
    \centering
    \includegraphics[width=\textwidth]{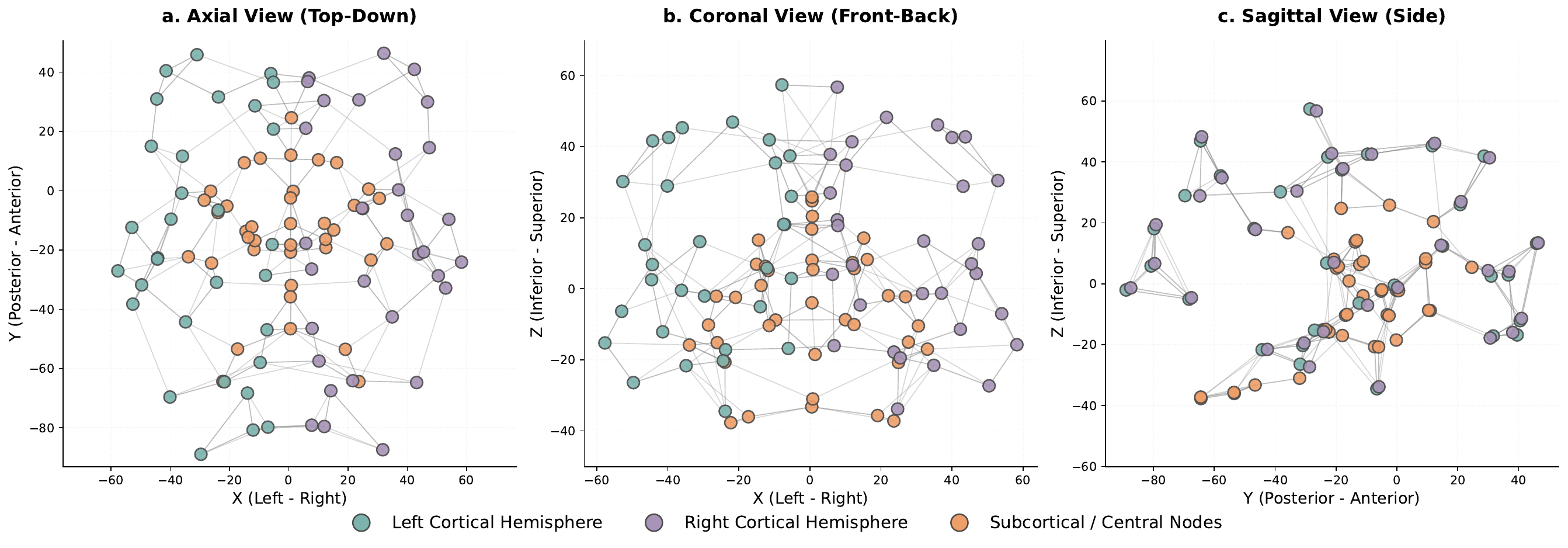}
    \caption{Orthographic projections of the population-level spatial graph embedded in standardized MNI space. The graph comprises 101 neuroanatomical nodes corresponding to regions from the DKT and ASEG atlases, color-coded by macro-anatomical division. Undirected edges represent K-nearest spatial neighbors ($K=3$) computed using Euclidean distance between regional centers of mass. The projections illustrate the spatial topology across the axial, coronal, and sagittal views, while the adjacency matrix highlights the sparse connectivity used to constrain ST-GNN message passing.}
    \label{fig:spatial_graph}
\end{figure*}  

\section{Methods}
\label{sec:methods}

\subsection{Dataset and Cognitive Phenotypes}

Data were obtained from the Healthy Brain Network (HBN), an open-access pediatric neurodevelopmental dataset \cite{alexander2017open}. Demographic, socioeconomic, cognitive, family, and structural MRI information were integrated for analysis. Four cognitive phenotypes were modeled independently: Dimensional Change Card Sort, Flanker Inhibitory Control and Attention, List Sorting Working Memory, and Pattern Comparison Processing Speed.

Age-uncorrected standardized NIH Toolbox scores were used as prediction targets to ensure that developmental variation in cognitive performance was represented in the outcome. The model covariates comprised age, sex, handedness, estimated total intracranial volume (eTIV), Parental education, Parental occupation, financial support status, and acquisition site. Target-specific cohort characteristics are summarized in Table~\ref{tab:demographics_imputed}.

\begin{table*}[t]
\centering
\caption{Family-aware stratified cross-validation fold statistics across cognitive targets}
\label{tab:fold_statistics}
\renewcommand{\arraystretch}{1.15}
\begin{tabular*}{\textwidth}{@{\extracolsep{\fill}} ll cccc @{}}
\toprule
\multicolumn{2}{l}{\multirow{2}{*}{\textbf{Data Partition}}} & \multicolumn{4}{c}{\textbf{Cognitive Target Distributions [N (Mean $\pm$ SD)]}} \\
\cmidrule{3-6}
& & \textbf{Card Sort} & \textbf{Flanker} & \textbf{List Sort} & \textbf{Processing Speed} \\
\midrule
\multirow{3}{*}{\textbf{Fold 1}} & Train & 1648 (87.26 $\pm$ 19.18) & 1647 (86.25 $\pm$ 15.53) & 1616 (92.64 $\pm$ 17.47) & 1630 (88.67 $\pm$ 19.78) \\
 & Val & 291 (87.17 $\pm$ 20.29) & 291 (85.69 $\pm$ 16.08) & 286 (91.22 $\pm$ 19.28) & 288 (87.74 $\pm$ 20.91) \\
 & Test & 485 (86.48 $\pm$ 20.05) & 484 (85.82 $\pm$ 16.19) & 476 (92.36 $\pm$ 18.18) & 480 (88.81 $\pm$ 20.20) \\
\addlinespace
\multirow{3}{*}{\textbf{Fold 2}} & Train & 1648 (87.17 $\pm$ 19.43) & 1646 (86.08 $\pm$ 15.48) & 1616 (92.33 $\pm$ 17.78) & 1630 (88.51 $\pm$ 19.76) \\
 & Val & 291 (86.05 $\pm$ 19.76) & 291 (85.41 $\pm$ 15.79) & 286 (92.80 $\pm$ 18.31) & 288 (89.93 $\pm$ 20.64) \\
 & Test & 485 (87.46 $\pm$ 19.52) & 485 (86.53 $\pm$ 16.52) & 476 (92.47 $\pm$ 17.76) & 480 (88.03 $\pm$ 20.41) \\
\addlinespace
\multirow{3}{*}{\textbf{Fold 3}} & Train & 1647 (86.88 $\pm$ 19.51) & 1647 (85.96 $\pm$ 15.93) & 1617 (92.73 $\pm$ 17.83) & 1631 (88.41 $\pm$ 20.08) \\
 & Val & 291 (89.17 $\pm$ 18.60) & 291 (85.28 $\pm$ 16.47) & 286 (90.77 $\pm$ 18.30) & 288 (88.86 $\pm$ 19.98) \\
 & Test & 486 (86.57 $\pm$ 19.87) & 484 (87.05 $\pm$ 14.51) & 475 (92.33 $\pm$ 17.54) & 479 (89.02 $\pm$ 19.78) \\
\addlinespace
\multirow{3}{*}{\textbf{Fold 4}} & Train & 1649 (86.76 $\pm$ 19.68) & 1647 (86.23 $\pm$ 15.57) & 1616 (92.39 $\pm$ 17.95) & 1630 (88.62 $\pm$ 20.05) \\
 & Val & 291 (88.87 $\pm$ 18.99) & 291 (86.22 $\pm$ 16.32) & 286 (91.20 $\pm$ 18.43) & 288 (88.65 $\pm$ 20.50) \\
 & Test & 484 (87.17 $\pm$ 19.11) & 484 (85.55 $\pm$ 15.92) & 476 (93.21 $\pm$ 17.07) & 480 (88.42 $\pm$ 19.55) \\
\addlinespace
\multirow{3}{*}{\textbf{Fold 5}} & Train & 1649 (87.00 $\pm$ 19.68) & 1646 (86.41 $\pm$ 15.70) & 1617 (92.61 $\pm$ 17.70) & 1631 (88.54 $\pm$ 20.07) \\
 & Val & 291 (86.42 $\pm$ 19.35) & 291 (85.26 $\pm$ 16.37) & 286 (92.47 $\pm$ 17.30) & 288 (88.77 $\pm$ 19.47) \\
 & Test & 484 (87.81 $\pm$ 18.90) & 485 (85.52 $\pm$ 15.44) & 475 (91.70 $\pm$ 18.62) & 479 (88.65 $\pm$ 20.10) \\
\bottomrule
\end{tabular*}

\vspace{4pt}
\raggedright
\footnotesize \textit{Note:} Subject counts denote the final complete-case cohort per cross-validation fold after filtering for target-specific missingness. Following a strict family-aware grouping strategy, siblings were guaranteed to remain within the same split (train, validation, or test) simultaneously, ensuring zero familial data leakage. Stratification via ten-quantile binning successfully stabilized target distributions across all partitions.
\end{table*}

\subsection{Structural MRI Processing}

T1-weighted MRI data were represented using FreeSurfer-derived regional morphometry. The anatomical representation contained 101 regions, comprising 62 bilateral cortical regions from the Desikan--Killiany--Tourville (DKT) atlas \cite{Klein2012} and 39 subcortical, ventricular, brainstem, and related structures from the ASEG atlas \cite{Fischl2002}.

FreeSurfer quality control was performed using hemispheric Euler numbers, a validated objective measure of topological complexity and structural image quality \cite{Rosen2018}. Participants were excluded when either hemisphere had an Euler number below $-217$ or when the required anatomical regions were unavailable. Of 3,736 subjects assessed during structural feature compilation, 450 failed the Euler-number criterion, and three were excluded because of missing anatomical data, leaving 3,283 participants with extracted structural features. Intersection with the curated phenotypic dataset yielded a graph-aligned cohort of 2,427 participants. Target-specific cohorts $(N\approx2,378-2,424)$ were finalized by excluding individuals missing the specific NIH Toolbox assessment.

For cortical regions, average thickness, surface area, and gray-matter volume were extracted. ASEG structures contributed to the regional volume. These measurements were organized into a unified $101 \times 3$ representation corresponding to volume, surface area, and thickness. For subcortical structures for which surface area or cortical thickness was not defined, the corresponding entries were set to zero to maintain a consistent tensor representation. While preserving tensor dimensions, this zero-filling creates a distinct numerical pattern of missingness that enables the model to inherently distinguish cortical from subcortical nodes.

\begin{table*}[t]
\centering
\caption{Neuroanatomical regions and associated MNI spatial coordinates used for template graph construction}
\label{tab:mni_coordinates}
\resizebox{\textwidth}{!}{
\begin{tabular}{l ccc ccc | l ccc ccc}
\toprule
\multirow{2}{*}{\textbf{Region Name}} & \multicolumn{3}{c}{\textbf{Left (MNI)}} & \multicolumn{3}{c|}{\textbf{Right (MNI)}} & \multirow{2}{*}{\textbf{Region Name}} & \multicolumn{3}{c}{\textbf{Left (MNI)}} & \multicolumn{3}{c}{\textbf{Right (MNI)}} \\
& \textbf{X} & \textbf{Y} & \textbf{Z} & \textbf{X} & \textbf{Y} & \textbf{Z} & & \textbf{X} & \textbf{Y} & \textbf{Z} & \textbf{X} & \textbf{Y} & \textbf{Z} \\
\midrule
\multicolumn{14}{l}{\textit{\textbf{Cortical Regions (DKT Atlas)}} \textit{[62 regions total: 31 Left, 31 Right]}} \\
\midrule
Caudal Anterior Cingulate & -5.20 & 20.79 & 26.02 & 5.73 & 21.07 & 26.96 & Pars Opercularis & -46.43 & 15.00 & 12.38 & 47.44 & 14.52 & 12.65 \\
Caudal Middle Frontal & -35.91 & 11.64 & 45.33 & 35.98 & 12.41 & 46.12 & Pars Orbitalis & -41.43 & 40.46 & -12.12 & 42.39 & 40.94 & -11.38 \\
Cuneus & -7.06 & -79.77 & 18.11 & 7.73 & -79.12 & 19.40 & Pars Triangularis & -44.56 & 30.95 & 2.54 & 46.83 & 29.95 & 4.32 \\
Entorhinal & -23.84 & -6.58 & -34.46 & 24.71 & -5.89 & -33.84 & Pericalcarine & -12.05 & -80.75 & 5.83 & 11.98 & -79.56 & 6.63 \\
Fusiform & -34.84 & -44.28 & -21.65 & 34.90 & -42.56 & -21.54 & Postcentral & -44.25 & -22.71 & 41.63 & 43.79 & -21.43 & 42.76 \\
Inferior Parietal & -40.12 & -69.61 & 28.93 & 43.15 & -64.72 & 28.86 & Posterior Cingulate & -5.63 & -18.13 & 37.40 & 5.74 & -17.73 & 37.83 \\
Inferior Temporal & -49.68 & -31.77 & -26.41 & 50.46 & -28.70 & -27.30 & Precentral & -39.77 & -9.60 & 42.57 & 40.04 & -8.30 & 42.56 \\
Insula & -36.15 & -0.83 & -0.44 & 37.06 & 0.27 & -1.24 & Precuneus & -9.68 & -57.95 & 35.44 & 10.20 & -57.47 & 34.81 \\
Isthmus Cingulate & -7.37 & -46.93 & 18.14 & 7.87 & -46.46 & 17.81 & Rostral Anterior Cingulate & -5.21 & 36.60 & 2.91 & 6.37 & 36.84 & 4.06 \\
Lateral Occipital & -29.66 & -88.94 & -2.03 & 31.72 & -87.41 & -1.34 & Rostral Middle Frontal & -31.00 & 45.89 & 13.27 & 32.08 & 46.38 & 13.44 \\
Lateral Orbitofrontal & -23.76 & 31.63 & -17.13 & 23.67 & 30.65 & -17.79 & Superior Frontal & -11.42 & 28.64 & 41.92 & 11.86 & 30.42 & 41.37 \\
Lingual & -13.96 & -68.28 & -5.05 & 14.18 & -67.49 & -4.57 & Superior Parietal & -21.76 & -64.48 & 46.92 & 21.58 & -64.11 & 48.24 \\
Medial Orbitofrontal & -6.07 & 39.49 & -16.79 & 6.82 & 38.05 & -16.00 & Superior Temporal & -52.98 & -12.40 & -6.33 & 54.01 & -9.65 & -7.03 \\
Middle Temporal & -57.70 & -27.05 & -15.24 & 58.29 & -24.09 & -15.73 & Supramarginal & -52.66 & -38.26 & 30.18 & 52.94 & -32.83 & 30.46 \\
Parahippocampal & -24.39 & -30.92 & -20.25 & 25.46 & -30.50 & -19.44 & Transverse Temporal & -44.34 & -23.06 & 6.82 & 45.51 & -20.67 & 7.02 \\
Paracentral & -7.86 & -28.57 & 57.38 & 7.72 & -26.45 & 56.75 &  &  &  &  &  &  &  \\
\midrule
\multicolumn{14}{l}{\textit{\textbf{Subcortical Regions (ASEG Atlas)}} \textit{[39 regions total: 14 Left, 14 Right, 11 Central]}} \\
\midrule
Lateral Ventricle & -14.46 & -13.69 & 13.71 & 15.18 & -13.24 & 14.26 & Choroid Plexus & -12.50 & -12.22 & 6.31 & 12.01 & -11.02 & 7.37 \\
Inferior Lateral Ventricle & -33.91 & -22.28 & -15.82 & 33.10 & -17.92 & -16.98 & 3rd Ventricle$^*$ & 0.58 & -11.12 & -3.94 & - & - & - \\
Cerebellum White Matter & -17.31 & -53.43 & -36.02 & 19.07 & -53.51 & -35.67 & 4th Ventricle$^*$ & 0.54 & -46.56 & -33.22 & - & - & - \\
Cerebellum Cortex & -22.27 & -64.37 & -37.67 & 23.65 & -64.42 & -37.20 & Brain Stem$^*$ & 0.83 & -31.95 & -31.00 & - & - & - \\
Thalamus & -11.75 & -19.88 & 5.17 & 12.49 & -19.23 & 5.66 & CSF$^*$ & 0.60 & -20.71 & 8.04 & - & - & - \\
Caudate & -15.03 & 9.44 & 6.91 & 16.16 & 9.46 & 8.24 & WM Hypointensities$^*$ & -13.71 & -15.71 & 0.93 & - & - & - \\
Putamen & -26.37 & -0.16 & -2.09 & 26.94 & 0.55 & -2.25 & Optic Chiasm$^*$ & 1.47 & -0.20 & -18.47 & - & - & - \\
Pallidum & -20.92 & -5.19 & -2.41 & 22.11 & -4.94 & -1.95 & CC Posterior$^*$ & 0.54 & -35.80 & 16.78 & - & - & - \\
Hippocampus & -26.09 & -24.43 & -15.15 & 27.74 & -23.38 & -14.99 & CC Mid Posterior$^*$ & 0.61 & -18.29 & 24.75 & - & - & - \\
Amygdala & -23.91 & -7.34 & -20.64 & 24.94 & -6.01 & -20.74 & CC Central$^*$ & 0.58 & -2.42 & 25.82 & - & - & - \\
Accumbens Area & -9.63 & 10.96 & -8.79 & 9.99 & 10.42 & -8.71 & CC Mid Anterior$^*$ & 0.70 & 12.00 & 20.41 & - & - & - \\
Ventral Diencephalon & -11.46 & -16.85 & -10.33 & 12.43 & -16.41 & -10.06 & CC Anterior$^*$ & 0.88 & 24.61 & 5.41 & - & - & - \\
Vessel & -28.55 & -3.20 & -10.18 & 30.56 & -2.58 & -10.44 &  &  &  &  &  &  &  \\
\bottomrule
\end{tabular}
}

\vspace{4pt}
\raggedright
\footnotesize \textbf{Note:} Coordinates are provided in MNI space, where X maps left (-) to right (+), Y maps posterior (-) to anterior (+), and Z maps inferior (-) to superior (+). Asterisks (*) denote unpaired midline structures. For conciseness in formatting, spatial coordinates for midline structures are reported in the left-hemisphere columns. CSF: cerebrospinal fluid; WM: white matter; CC: corpus callosum.
\end{table*}

\subsection{Anatomical Coordinates and Spatial Graph}

A standard anatomical reference was established using the \textit{fsaverage} template \cite{Fischl1999} via MNE-Python \cite{Gramfort2013}. The center of mass for each anatomical region in the template volume was converted into MNI coordinates via the relevant affine transformation. The complete set of 101 anatomical regions and their coordinates is provided in Table~\ref{tab:mni_coordinates}.

A population-level spatial graph was then constructed from these coordinates \cite{parisot2018disease}. For each anatomical node, the three nearest neighbors were identified using Euclidean distance to form a localized spatial topology (see Figure \ref{fig:spatial_graph}). The resulting adjacency matrix was defined as:

\begin{equation}
A_{ij} =
\begin{cases}
1, & \text{if } j \in \mathcal{N}_{K}(i) \lor i \in \mathcal{N}_{K}(j) \\
0, & \text{otherwise}
\end{cases}
\qquad (K=3),
\label{eq:spatial_adjacency}
\end{equation}

where $\mathcal{N}_{K}(i)$ denotes the $K$ nearest spatial neighbors of node $i$. Reciprocal edges were added to obtain an undirected topology, and self-connections were included during message passing. The resulting graph was fixed across participants, while the node features varied between subjects. Because the graph is constructed from template-space coordinates rather than subject-specific connectivity measurements, the resulting edges should be interpreted as spatial proximity relationships in standardized MNI space rather than as direct measures of anatomical or functional connectivity. To validate the necessity of localized spatial constraints, the baseline topology ($K=3$) was systematically evaluated against broader spatial neighborhoods ($K \in \{5, 10, 15\}$) and a fully connected (all-to-all) graph configuration. This allowed us to explicitly test whether predictive performance stems from sparse geometric routing or simply the addition of non-linear parameterizations. The main ST-GNN configuration was fixed a priori at $K=3$; the alternative $K$ values were evaluated as post hoc architectural ablations to characterize the sensitivity of predictive performance to neighborhood size rather than to select the primary model using the held-out test results.

\subsection{Family-Aware Data Partitioning and Preprocessing}


To prevent leakage of familial data during cross-validation, participants were explicitly grouped using sibling-linkage identifiers. Because only approximately 4\% of the cohort possessed a sibling, applying stringent multimodal inclusion criteria naturally excluded at least one member of every sibling pair, ensuring cohorts of strictly independent participants. Consequently, the final target-specific cohorts consisted exclusively of independent participants, rendering the family-aware partitions functionally equivalent to strictly independent stratification. For each cognitive target, subjects missing that specific assessment were excluded, and the continuous target distribution was divided into 10 quantile-based bins for stratification.

A five-fold cross-validation method was employed using the constructed Family IDs as the grouping variable with a fixed random seed of 42. Within each training–validation pool, a group-level split reserved 15\% for validation. This strategy was applied independently for each cognitive target, guaranteeing that the held-out test partition contained no participants from families represented in the corresponding training or validation partitions. The resulting fold statistics are reported in Table~\ref{tab:fold_statistics}.

All preprocessing parameters were estimated strictly using training data to prevent information leakage. Continuous covariates were median-imputed and standardized, while categorical covariates were mode-imputed and one-hot encoded. Target standardization for neural-network experiments was similarly derived solely from the training partition and inverted before the final evaluation metric calculation.

\subsection{Conventional Machine-Learning Baselines}
Ridge, Lasso, and Elastic Net regression were used as conventional machine-learning baselines. Hyperparameters were optimized using the predefined training/validation partitions.

Two feature configurations were evaluated: \textit{Covariates Only} and \textit{Covariates + Brain}. The latter incorporated the 303-dimensional structural representation together with the same covariates. This comparison quantified the predictive contribution of structural MRI beyond demographic and socioeconomic information.

For each fold, the selected model was refitted on the training partition and evaluated on the held-out test partition.

\subsection{Multilayer Perceptron (MLP) Baseline}

An MLP was implemented as a topology-free nonlinear baseline, using the same structural input as the ST-GNN. The $101 \times 3$ representation was flattened into a 303-dimensional vector:

\begin{equation}
\mathbf{x}^{\mathrm{flat}}_i = \operatorname{vec}(\mathbf{X}_i),
\label{eq:mlp_flatten}
\end{equation}

and processed through two fully connected layers with 256 hidden units, batch normalization, ReLU activation, and dropout ($p=0.3$). Covariates were processed through a separate 16-dimensional branch and fused with the structural representation through a 32-unit fully connected layer, followed by a single regression output.

The MLP used the same training and evaluation protocol as the ST-GNN, providing a topology-free nonlinear baseline with the same structural inputs and covariates.

\subsection{Spatial-Topological Graph Neural Network}

The proposed ST-GNN operates directly on the 101-node spatial representation. Each node was augmented with a trainable 16-dimensional node-identity embedding before graph processing.

Two graph-attention layers with 256 hidden channels were used. For node representations $\mathbf{h}_i$, query, key, and value vectors were obtained through learned linear projections. Attention was calculated using scaled dot-product attention restricted to the predefined spatial topology:

\begin{equation}
\alpha_{ij} =
\frac{
\exp\left(\mathbf{q}_i^{\top}\mathbf{k}_j/\sqrt{d_k}\right)
}{
\sum_{k\in\mathcal{N}^{+}(i)}
\exp\left(\mathbf{q}_i^{\top}\mathbf{k}_k/\sqrt{d_k}\right)
},
\label{eq:graph_attention}
\end{equation}

where $d_k$ is the key dimensionality and $\mathcal{N}^{+}(i)$ contains the spatial neighbors and self-connection. The resulting weighted value representations were passed through ELU activation. Batch normalization and dropout were applied after graph processing, and the second graph-attention layer incorporated a residual connection.

The resulting node embeddings were aggregated using learned global attention pooling to obtain a whole-brain structural representation. Covariates were processed through an independent 16-dimensional branch and introduced only after graph propagation and pooling. This late-fusion architectural design was explicitly chosen to prevent non-spatial systemic variables from artificially modifying the localized geometric message-passing steps, ensuring that the intermediate graph embeddings learned a strictly anatomical representation. The final prediction was therefore:

\begin{equation}
\hat{y}_i =
\operatorname{MLP}_{\mathrm{fusion}}
\left(
\mathbf{h}_{\mathrm{graph},i}
\parallel
\operatorname{MLP}_{\mathrm{cov}}(\mathbf{c}_i)
\right),
\label{eq:late_fusion}
\end{equation}

where $\parallel$ denotes concatenation. The fusion network consisted of a 32-unit hidden layer with ReLU activation, dropout, and a single regression output.

\subsection{Training and Comparative Experiments}

The ST-GNN and MLP were trained independently for each target and fold using Adam, MSE loss, a learning rate of $10^{-3}$, weight decay of $10^{-3}$, a batch size of 128, and a maximum of 100 epochs. Early stopping with a patience of 15 epochs was based on validation loss. The model with the lowest validation loss was restored before test evaluation. A fixed random seed of 42 was used for reproducibility.

The comparative design progressively evaluated the contribution of structural information and spatial modeling. Conventional ML models compared \textit{Covariates Only} with \textit{Covariates + Brain}, while the MLP and ST-GNN received identical structural and covariate inputs but differed in representation: the MLP flattened the structural tensor, whereas the ST-GNN retained its spatial topology. All competing models used the same target-specific held-out test partitions.

To retain between-region differences in morphometric magnitude, each structural feature channel (volume, surface area, and cortical thickness) was standardized globally using the non-zero training values pooled across all anatomical regions and training subjects within each fold. The resulting training-set mean and standard deviation were then applied unchanged to the corresponding validation and test partitions, while zero entries representing unavailable measurements were left untransformed.

\subsection{Explainable Artificial Intelligence}

Model attribution was examined using Integrated Gradients (IG) \cite{sundararajan2017axiomatic}. Attribution was computed on the held-out test subjects using a zero baseline and 50 integration steps. The resulting attributions were interpreted as model-level contribution scores rather than causal effects or direct evidence of biological mechanisms. For feature $x_i$ relative to baseline $x_i'$, IG was defined as:

\begin{equation}
\operatorname{IG}_i(x) =
(x_i-x_i')
\int_{0}^{1}
\frac{
\partial F\left(x'+\alpha(x-x')\right)
}{
\partial x_i
}
\,d\alpha,
\label{eq:integrated_gradients}
\end{equation}

where $F$ denotes the trained prediction function.

Absolute structural attributions were averaged across subjects within each fold and subsequently across folds. Volume, surface area, and thickness contributions were retained separately for each anatomical region, while their combined magnitude was used as regional importance. A whole-brain structural importance score was also calculated by aggregating absolute attributions across all nodes and channels. Covariate attributions were aggregated in a similar manner to enable comparison between structural morphology and demographic or socioeconomic factors.

\subsection{Topological Attention Analysis}

Attention maps from both graph-attention layers and the global pooling mechanism were extracted during test inference. For anatomical node $i$, the attention-based routing score was calculated as:

\begin{equation}
\operatorname{Hub}(i) =
\frac{1}{H}
\sum_{h=1}^{H}
\sum_{j=1}^{101}
\alpha_{ji}^{(h)},
\label{eq:hub_score}
\end{equation}

where $H$ denotes the attention dimensions and $\alpha_{ji}^{(h)}$ represents the attention assigned to node $i$ from node $j$. Attention scores were aggregated across subjects and folds and interpreted alongside IG-derived regional importance.

\subsection{Evaluation Metrics}

Performance was evaluated using RMSE, MAE, Pearson correlation coefficient ($r$), and coefficient of determination ($R^2$). All metrics were calculated on the held-out test partitions after inverse transformation to the original target scale. Results were summarized across the five folds as mean $\pm$ standard deviation. 

\begin{table*}[htbp]
\centering
\caption{Baseline predictive performance using demographic and socioeconomic covariates only}
\label{tab:covariates_only}
\renewcommand{\arraystretch}{1.15}
\begin{tabular*}{\textwidth}{@{\extracolsep{\fill}} l l cccc @{}}
\toprule
\textbf{Cognitive Target} & \textbf{Linear Model} & $\mathbf{R^2}$ & $\mathbf{r}$ & \textbf{RMSE} & \textbf{MAE} \\
\midrule
\multirow{3}{*}{\textbf{Card Sort}} 
 & Ridge & 0.3621 $\pm$ 0.0159 & 0.6041 $\pm$ 0.0132 & 15.55 $\pm$ 0.48 & 11.71 $\pm$ 0.36 \\
 & Lasso & \textbf{0.3625 $\pm$ 0.0161} & \textbf{0.6041 $\pm$ 0.0133} & \textbf{15.55 $\pm$ 0.48} & 11.71 $\pm$ 0.38 \\
 & ElasticNet & 0.3616 $\pm$ 0.0159 & 0.6041 $\pm$ 0.0134 & 15.56 $\pm$ 0.49 & \textbf{11.70 $\pm$ 0.38} \\
\addlinespace
\multirow{3}{*}{\textbf{Flanker}} 
 & Ridge & \textbf{0.3081 $\pm$ 0.0215} & 0.5594 $\pm$ 0.0225 & \textbf{13.06 $\pm$ 0.67} & 9.84 $\pm$ 0.49 \\
 & Lasso & 0.3078 $\pm$ 0.0210 & \textbf{0.5595 $\pm$ 0.0227} & 13.06 $\pm$ 0.67 & \textbf{9.83 $\pm$ 0.49} \\
 & ElasticNet & 0.3081 $\pm$ 0.0215 & 0.5594 $\pm$ 0.0225 & 13.06 $\pm$ 0.67 & 9.84 $\pm$ 0.49 \\
\addlinespace
\multirow{3}{*}{\textbf{List Sort}} 
 & Ridge & \textbf{0.3296 $\pm$ 0.0293} & 0.5774 $\pm$ 0.0264 & \textbf{14.58 $\pm$ 0.34} & \textbf{11.35 $\pm$ 0.24} \\
 & Lasso & 0.3270 $\pm$ 0.0317 & 0.5748 $\pm$ 0.0280 & 14.61 $\pm$ 0.30 & 11.39 $\pm$ 0.22 \\
 & ElasticNet & 0.3295 $\pm$ 0.0290 & \textbf{0.5776 $\pm$ 0.0263} & 14.58 $\pm$ 0.35 & 11.35 $\pm$ 0.24 \\
\addlinespace
\multirow{3}{*}{\textbf{Processing Speed}} 
 & Ridge & 0.4021 $\pm$ 0.0272 & 0.6365 $\pm$ 0.0207 & 15.46 $\pm$ 0.56 & 12.25 $\pm$ 0.23 \\
 & Lasso & 0.4020 $\pm$ 0.0287 & 0.6360 $\pm$ 0.0213 & 15.46 $\pm$ 0.58 & \textbf{12.23 $\pm$ 0.24} \\
 & ElasticNet & \textbf{0.4025 $\pm$ 0.0274} & \textbf{0.6369 $\pm$ 0.0210} & \textbf{15.45 $\pm$ 0.57} & 12.24 $\pm$ 0.23 \\
\bottomrule
\end{tabular*}

\vspace{4pt}
\raggedright
\footnotesize \textit{Note:} Bold values indicate the best-performing model for each metric within each cognitive target.
\end{table*}

\begin{table*}[t]
\footnotesize
\centering
\setlength{\tabcolsep}{4pt}
\caption{Ablation analysis of ST-GNN topology, feature scaling, and covariate fusion}
\label{tab:stgnn_ablations}
\renewcommand{\arraystretch}{1.15}
\begin{tabular*}{\textwidth}{@{\extracolsep{\fill}} l l cccc @{}}
\toprule
\textbf{Cognitive Target} & \textbf{ST-GNN Configuration} & $\mathbf{R^2}$ & $\mathbf{r}$ & \textbf{RMSE} & \textbf{MAE} \\
\midrule
\multirow{9}{*}{\textbf{Card Sort}} 
 & \textit{Graph Connectivity} \\
 & \quad -- Spatial ($K=3$) & 0.4433 $\pm$ 0.0183 & \textbf{0.6677 $\pm$ 0.0151} & 14.5239 $\pm$ 0.3818 & \textbf{11.0068 $\pm$ 0.3326} \\
 & \quad -- Spatial ($K=5$) & 0.4419 $\pm$ 0.0227 & 0.6676 $\pm$ 0.0173 & 14.5399 $\pm$ 0.3273 & 11.0116 $\pm$ 0.2484 \\
 & \quad -- Spatial ($K=10$) & \textbf{0.4443 $\pm$ 0.0206} & 0.6674 $\pm$ 0.0160 & \textbf{14.5106 $\pm$ 0.3928} & 11.0278 $\pm$ 0.2701 \\
 & \quad -- Spatial ($K=15$) & 0.4426 $\pm$ 0.0226 & 0.6661 $\pm$ 0.0175 & 14.5328 $\pm$ 0.4266 & 11.0425 $\pm$ 0.3077 \\
 & \quad -- Fully Connected & 0.4356 $\pm$ 0.0179 & 0.6607 $\pm$ 0.0135 & 14.6256 $\pm$ 0.4224 & 11.1633 $\pm$ 0.3507 \\
 & \quad -- Randomized & 0.4334 $\pm$ 0.0231 & 0.6612 $\pm$ 0.0173 & 14.6476 $\pm$ 0.2327 & 11.2163 $\pm$ 0.2186 \\
 \addlinespace
 & \textit{Feature Scaling \& Systemic Fusion} \\
 & \quad -- Node-Wise ($K=3$) & 0.4254 $\pm$ 0.0276 & 0.6539 $\pm$ 0.0217 & 14.7554 $\pm$ 0.5124 & 11.1321 $\pm$ 0.4065 \\
 & \quad -- No Covariates ($K=3$) & 0.2124 $\pm$ 0.0245 & 0.4648 $\pm$ 0.0231 & 17.2815 $\pm$ 0.6087 & 13.1813 $\pm$ 0.5521 \\
\midrule
\multirow{9}{*}{\textbf{Flanker}} 
 & \textit{Graph Connectivity} \\
 & \quad -- Spatial ($K=3$) & 0.3963 $\pm$ 0.0264 & 0.6363 $\pm$ 0.0161 & 12.1881 $\pm$ 0.4771 & 9.3178 $\pm$ 0.3146 \\
 & \quad -- Spatial ($K=5$) & \textbf{0.3996 $\pm$ 0.0179} & \textbf{0.6364 $\pm$ 0.0135} & \textbf{12.1612 $\pm$ 0.5612} & 9.2448 $\pm$ 0.3987 \\
 & \quad -- Spatial ($K=10$) & 0.3895 $\pm$ 0.0282 & 0.6290 $\pm$ 0.0204 & 12.2607 $\pm$ 0.5866 & 9.3738 $\pm$ 0.4488 \\
 & \quad -- Spatial ($K=15$) & 0.3964 $\pm$ 0.0252 & 0.6350 $\pm$ 0.0187 & 12.1940 $\pm$ 0.6215 & \textbf{9.2434 $\pm$ 0.4240} \\
 & \quad -- Fully Connected & 0.3930 $\pm$ 0.0167 & 0.6307 $\pm$ 0.0127 & 12.2295 $\pm$ 0.5757 & 9.2747 $\pm$ 0.4043 \\
 & \quad -- Randomized & 0.3843 $\pm$ 0.0267 & 0.6314 $\pm$ 0.0140 & 12.3060 $\pm$ 0.4100 & 9.3288 $\pm$ 0.2244 \\
 \addlinespace
 & \textit{Feature Scaling \& Systemic Fusion} \\
 & \quad -- Node-Wise ($K=3$) & 0.3922 $\pm$ 0.0201 & 0.6300 $\pm$ 0.0170 & 12.2394 $\pm$ 0.6469 & 9.2643 $\pm$ 0.4742 \\
 & \quad -- No Covariates ($K=3$) & 0.1908 $\pm$ 0.0318 & 0.4447 $\pm$ 0.0319 & 14.1114 $\pm$ 0.5214 & 10.7114 $\pm$ 0.3487 \\
\midrule
\multirow{9}{*}{\textbf{List Sort}} 
 & \textit{Graph Connectivity} \\
 & \quad -- Spatial ($K=3$) & \textbf{0.4083 $\pm$ 0.0255} & \textbf{0.6446 $\pm$ 0.0227} & \textbf{13.6971 $\pm$ 0.3023} & \textbf{10.5810 $\pm$ 0.2313} \\
 & \quad -- Spatial ($K=5$) & 0.4012 $\pm$ 0.0213 & 0.6401 $\pm$ 0.0164 & 13.7826 $\pm$ 0.3757 & 10.6523 $\pm$ 0.2728 \\
 & \quad -- Spatial ($K=10$) & 0.3972 $\pm$ 0.0174 & 0.6402 $\pm$ 0.0160 & 13.8277 $\pm$ 0.2839 & 10.7148 $\pm$ 0.2249 \\
 & \quad -- Spatial ($K=15$) & 0.3986 $\pm$ 0.0236 & 0.6371 $\pm$ 0.0168 & 13.8097 $\pm$ 0.2946 & 10.6453 $\pm$ 0.2465 \\
 & \quad -- Fully Connected & 0.3939 $\pm$ 0.0193 & 0.6329 $\pm$ 0.0189 & 13.8647 $\pm$ 0.2978 & 10.6829 $\pm$ 0.2037 \\
 & \quad -- Randomized & 0.3942 $\pm$ 0.0260 & 0.6361 $\pm$ 0.0233 & 13.8609 $\pm$ 0.3817 & 10.7018 $\pm$ 0.2630 \\
 \addlinespace
 & \textit{Feature Scaling \& Systemic Fusion} \\
 & \quad -- Node-Wise ($K=3$) & 0.3764 $\pm$ 0.0216 & 0.6207 $\pm$ 0.0128 & 14.0680 $\pm$ 0.4927 & 10.7484 $\pm$ 0.3251 \\
 & \quad -- No Covariates ($K=3$) & 0.2137 $\pm$ 0.0313 & 0.4699 $\pm$ 0.0337 & 15.7926 $\pm$ 0.4610 & 12.1743 $\pm$ 0.3159 \\
\midrule
\multirow{9}{*}{\textbf{Processing Speed}} 
 & \textit{Graph Connectivity} \\
 & \quad -- Spatial ($K=3$) & 0.4165 $\pm$ 0.0210 & 0.6481 $\pm$ 0.0135 & 15.2705 $\pm$ 0.4931 & \textbf{12.1278 $\pm$ 0.2930} \\
 & \quad -- Spatial ($K=5$) & \textbf{0.4185 $\pm$ 0.0214} & \textbf{0.6493 $\pm$ 0.0152} & \textbf{15.2454 $\pm$ 0.5077} & 12.1325 $\pm$ 0.3189 \\
 & \quad -- Spatial ($K=10$) & 0.4171 $\pm$ 0.0174 & 0.6476 $\pm$ 0.0142 & 15.2630 $\pm$ 0.4497 & 12.1761 $\pm$ 0.3005 \\
 & \quad -- Spatial ($K=15$) & 0.4132 $\pm$ 0.0180 & 0.6460 $\pm$ 0.0138 & 15.3142 $\pm$ 0.4458 & 12.1955 $\pm$ 0.3050 \\
 & \quad -- Fully Connected & 0.4161 $\pm$ 0.0225 & 0.6461 $\pm$ 0.0173 & 15.2760 $\pm$ 0.5209 & 12.1885 $\pm$ 0.3824 \\
 & \quad -- Randomized & 0.4125 $\pm$ 0.0141 & 0.6460 $\pm$ 0.0104 & 15.3224 $\pm$ 0.3482 & 12.1477 $\pm$ 0.1852 \\
 \addlinespace
 & \textit{Feature Scaling \& Systemic Fusion} \\
 & \quad -- Node-Wise ($K=3$) & 0.3496 $\pm$ 0.1053 & 0.6132 $\pm$ 0.0563 & 16.0799 $\pm$ 1.3544 & 12.3781 $\pm$ 0.4254 \\
 & \quad -- No Covariates ($K=3$) & 0.2415 $\pm$ 0.0432 & 0.4994 $\pm$ 0.0430 & 17.4027 $\pm$ 0.5755 & 13.9119 $\pm$ 0.6145 \\
\bottomrule
\end{tabular*}

\vspace{4pt}
\raggedright
\footnotesize \textit{Note:} Bold values indicate the best-performing configuration for each metric within each cognitive target[cite: 3]. To strictly isolate the impact of graph architecture, all configurations incorporate demographic covariates via late fusion and apply global feature scaling, except for the explicit ``Node-Wise'' and ``No Covariates'' ablation rows[cite: 3].
\end{table*}

\begin{table*}[t]
\centering
\caption{Predictive performance of baseline models using unstructured (flattened) structural morphometry and covariates}
\label{tab:flattened_baselines}
\renewcommand{\arraystretch}{1.15}
\begin{tabular*}{\textwidth}{@{\extracolsep{\fill}} l l cccc @{}}
\toprule
\textbf{Cognitive Target} & \textbf{Model} & $\mathbf{R^2}$ & $\mathbf{r}$ & \textbf{RMSE} & \textbf{MAE} \\
\midrule
\multirow{4}{*}{\textbf{Card Sort}} 
 & Ridge & 0.3719 $\pm$ 0.0205 & 0.6117 $\pm$ 0.0179 & 15.4300 $\pm$ 0.4900 & 11.7000 $\pm$ 0.3300 \\
 & Lasso & 0.3720 $\pm$ 0.0178 & 0.6119 $\pm$ 0.0153 & 15.4300 $\pm$ 0.4700 & \textbf{11.6700 $\pm$ 0.3200} \\
 & ElasticNet & \textbf{0.3726 $\pm$ 0.0183} & \textbf{0.6128 $\pm$ 0.0167} & \textbf{15.4200 $\pm$ 0.4800} & \textbf{11.6700 $\pm$ 0.3400} \\
 & MLP & 0.2774 $\pm$ 0.0294 & 0.5483 $\pm$ 0.0309 & 16.5502 $\pm$ 0.5904 & 12.4426 $\pm$ 0.6814 \\
\midrule
\multirow{4}{*}{\textbf{Flanker}} 
 & Ridge & \textbf{0.3212 $\pm$ 0.0200} & \textbf{0.5708 $\pm$ 0.0181} & \textbf{12.9300 $\pm$ 0.5600} & \textbf{9.7400 $\pm$ 0.4200} \\
 & Lasso & 0.3120 $\pm$ 0.0340 & 0.5660 $\pm$ 0.0249 & 13.0100 $\pm$ 0.4900 & 9.8100 $\pm$ 0.3600 \\
 & ElasticNet & 0.3142 $\pm$ 0.0301 & 0.5683 $\pm$ 0.0233 & 12.9900 $\pm$ 0.4800 & 9.8000 $\pm$ 0.3300 \\
 & MLP & 0.2357 $\pm$ 0.0247 & 0.5096 $\pm$ 0.0120 & 13.7159 $\pm$ 0.4947 & 10.3817 $\pm$ 0.3402 \\
\midrule
\multirow{4}{*}{\textbf{List Sort}} 
 & Ridge & 0.3410 $\pm$ 0.0361 & 0.5906 $\pm$ 0.0273 & 14.4600 $\pm$ 0.5600 & 11.1700 $\pm$ 0.4000 \\
 & Lasso & 0.3388 $\pm$ 0.0370 & 0.5866 $\pm$ 0.0307 & 14.4800 $\pm$ 0.4000 & 11.2500 $\pm$ 0.3400 \\
 & ElasticNet & \textbf{0.3489 $\pm$ 0.0317} & \textbf{0.5930 $\pm$ 0.0271} & \textbf{14.3700 $\pm$ 0.4000} & \textbf{11.1100 $\pm$ 0.2900} \\
 & MLP & 0.2760 $\pm$ 0.0312 & 0.5444 $\pm$ 0.0170 & 15.1519 $\pm$ 0.3740 & 11.7319 $\pm$ 0.2501 \\
\midrule
\multirow{4}{*}{\textbf{Processing Speed}} 
 & Ridge & 0.3952 $\pm$ 0.0230 & 0.6296 $\pm$ 0.0176 & 15.5500 $\pm$ 0.5100 & 12.3700 $\pm$ 0.2600 \\
 & Lasso & \textbf{0.3996 $\pm$ 0.0271} & \textbf{0.6336 $\pm$ 0.0202} & \textbf{15.4900 $\pm$ 0.5600} & \textbf{12.2800 $\pm$ 0.2500} \\
 & ElasticNet & 0.3947 $\pm$ 0.0255 & 0.6329 $\pm$ 0.0189 & 15.5500 $\pm$ 0.5500 & 12.3400 $\pm$ 0.2700 \\
 & MLP & 0.2886 $\pm$ 0.0554 & 0.5652 $\pm$ 0.0202 & 16.8516 $\pm$ 0.7946 & 13.4412 $\pm$ 0.5518 \\
\bottomrule
\end{tabular*}

\vspace{4pt}
\raggedright
\footnotesize \textit{Note:} Bold values indicate the best-performing model for each metric within each cognitive target.
\end{table*}

\begin{figure*}[htbp]
    \centering
    \includegraphics[width=\textwidth]{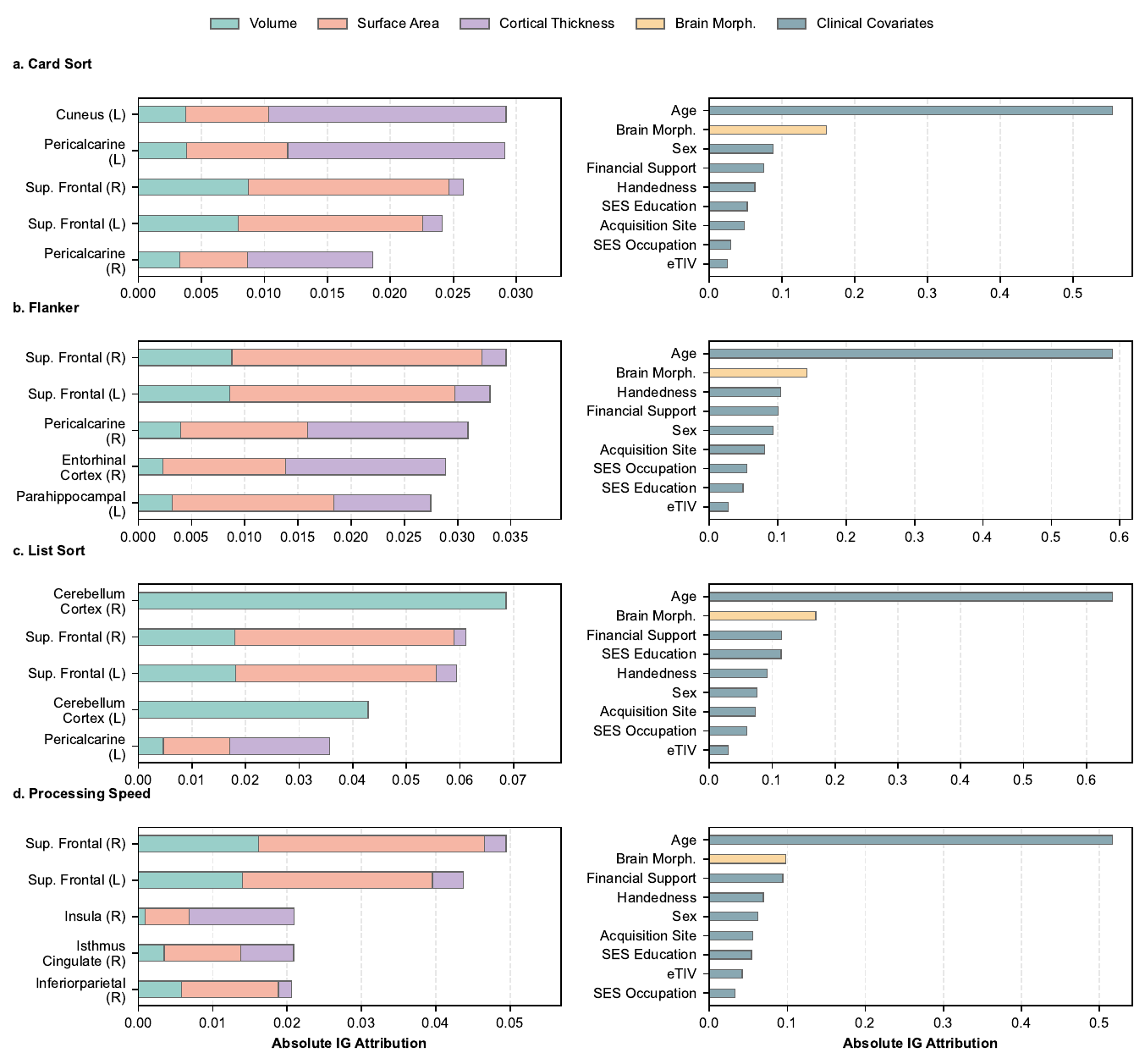}
    \caption{Integrated Gradients attribution analysis across four cognitive targets: (a) dimensional change card sort, (b) flanker inhibitory control and attention, (c) list sorting, and (d) processing speed. The left panels show the five morphological brain regions with the highest mean absolute attribution for each target, with attribution decomposed into volume, surface area, and cortical thickness components. The right panels compare the aggregate attribution of structural brain morphology with that of demographic, socioeconomic, and acquisition-related covariates.}
    \label{fig:xai_attribution}
\end{figure*}

\begin{figure*}[htbp]
    \centering
    \includegraphics[width=\textwidth]{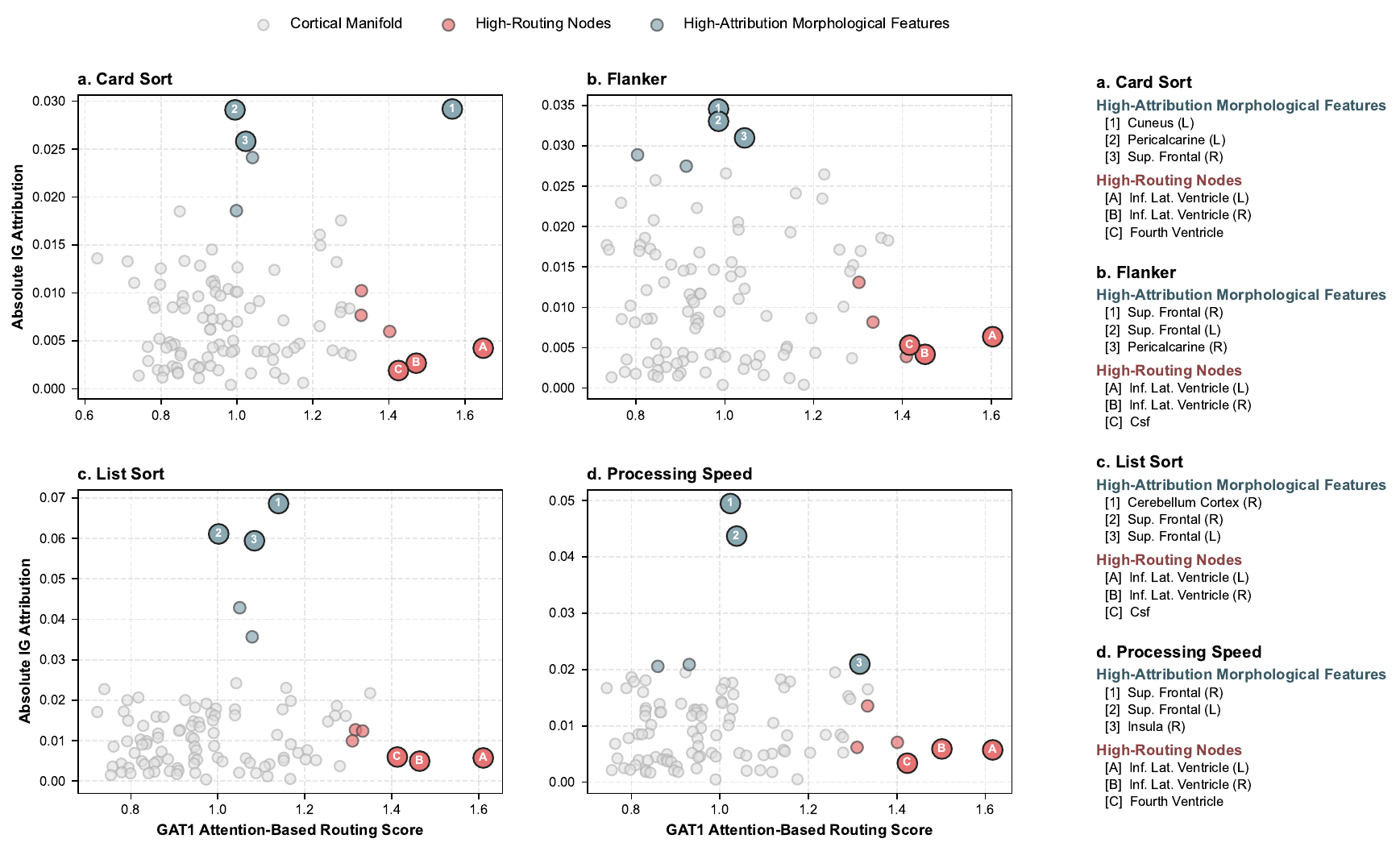}
    \caption{Relationship between attention-based routing and model attribution across the four cognitive targets. The scatter plots show spatial attention-based routing scores from the first graph-attention layer against absolute Integrated Gradients attribution. The plots illustrate that nodes with relatively high routing scores do not necessarily receive high predictive attribution. The high-routing, low-attribution nodes include centrally positioned fluid-space structures, whereas the high-attribution features include morphological regions identified by the model for each cognitive target.}
    \label{fig:gat_vs_ig_scatter}
\end{figure*}

\section{Results and Discussion}
\label{sec:results_discussion}

Accurate prediction of cognitive phenotypes in pediatric cohorts depends on separating confounded biological development processes from systemic environmental factors. By systematically evaluating conventional linear models, topology-free deep learning, and spatial graph ablations, we find that structural brain morphometry provides predictive information for cognitive performance, with the magnitude of the predictive benefit depending on the representation and spatial constraints used by the model.

\subsection{Systemic Baselines and the Limitations of Unstructured Morphology}
To establish a rigorous predictive floor, cognitive targets were initially modeled using only demographic and socioeconomic covariates. As reported in Table \ref{tab:covariates_only}, systemic factors consistently accounted for a considerable portion of cognitive variance, yielding the highest predictability for Processing Speed ($R^2 = 0.4025$) and the lowest for Flanker ($R^2 = 0.3081$). These results indicate that demographic and socioeconomic factors account for a substantial proportion of the observed variance in pediatric cognitive performance \cite{weintraub2013pediatric}.

A fundamental challenge in computational neuroimaging is successfully integrating high-dimensional structural data to improve upon these systemic baselines. When 303 regional morphological features were flattened into unstructured tabular inputs (Table \ref{tab:flattened_baselines}), they provided negligible or even negative gains. Linear models improved the variance explained by less than $\Delta R^2 = 0.02$ compared to covariate-only baselines. More critically, the topology-free MLP baseline suffered severe performance degradation across all tasks (e.g., Flanker RMSE increased to 13.71, $R^2 = 0.2357$). This result suggests that, for the present representation and sample, the topology-free MLP does not effectively exploit the structured relationships among regional morphometric features.

\subsection{The Necessity of Spatial Inductive Biases and Topological Sparsity}
The proposed ST-GNN overcomes the limitations of unstructured data by embedding regional morphometry into a localized geometric graph mapped to standardized MNI space. However, as demonstrated in the extensive ablation study (Table \ref{tab:stgnn_ablations}), simply mapping features to a graph is not inherently sufficient; the connections benefit from being constrained by spatial proximity.

When structural topology was fully connected (all-to-all message passing), performance consistently degraded across all metrics relative to the sparse configurations. For example, in the List Sort task, the fully connected graph yielded higher absolute errors (RMSE = 13.8647) and lower correlation ($r = 0.6329$) than its sparse spatial counterparts. This performance reduction suggests that unrestricted message passing may dilute localized morphometric information. By forcing every region to communicate directly with every other region, the network blends distinct localized morphometric signals until their task-specific predictive utility is washed out. To further validate this, the structural topology was ablated by randomly shuffling the adjacency matrix while preserving the exact node features and degree distribution. This Randomized Topology consistently reduced predictive accuracy, supporting the hypothesis that the spatial arrangement encoded by the proposed graph contributes to predictive performance beyond the node features and degree distribution alone.

Sparse spatial neighborhoods provided a useful architectural regularization in the present experiments, although the preferred neighborhood size varied across cognitive domains. For tasks involving rapid visual evaluation and cognitive flexibility---such as Processing Speed and Flanker---the $K=5$ configuration yielded the highest $R^2$ for Flanker and Processing Speed in the reported ablation results. In contrast, the $K=3$ configuration yielded the lowest RMSE for the working memory (List Sort) task. Across the alternative neighborhood sizes, the best-performing configuration varied by cognitive target and evaluation metric, indicating that predictive performance was relatively stable across the tested sparse graph densities rather than dominated by a single neighborhood size. Overall, sparse spatial configurations generally achieved better predictive performance than fully connected and randomized topologies, and outperformed the topology-free MLP baseline across the evaluated cognitive targets.

\subsection{Retaining Between-Region Scale Differences via Global Normalization}
Beyond graph topology, we observed a critical inductive bias regarding feature distribution. As detailed in Table \ref{tab:stgnn_ablations}, node-wise feature scaling---standardizing volume, area, and thickness independently for each anatomical region---substantially degraded the ST-GNN's predictive performance across all tasks. For instance, in the Processing Speed domain, node-wise scaling collapsed the explained variance ($R^2 = 0.3496 \pm 0.1053$) while introducing massive instability (RMSE = 16.0799) compared to the proposed global scaling strategy ($R^2 = 0.4165$, RMSE = 15.2705). 

Node-wise scaling artificially flattens the anatomical hierarchy; it forces the network to treat minor structural variance in small subcortical nuclei with the same mathematical weight as massive volumetric variance in the cortical lobes. Conversely, global scaling across the entire cortex preserves the brain's absolute morphometric scale. The observed advantage of global scaling suggests that retaining between-region differences in morphometric magnitude is beneficial for the present graph-based representation. While retaining between-region scale differences provides a useful inductive bias for spatial graph learning, it also enables the model to exploit absolute size differences that strongly correlate with systemic covariates such as age, sex, and total intracranial volume.

\subsection{Synergy of Structure and Systemic Development}
An ablation isolating structural features entirely (Morphology Only, Table \ref{tab:stgnn_ablations}) yielded substantially lower explained variance ($R^2 \approx 0.19$ to $0.24$) than demographic covariates alone. This biologically expected result affirms that regional brain structure alone cannot holistically account for cognitive variance driven by age, education, and socioeconomic environments \cite{snoek2019confounds}. The ST-GNN therefore provides additional predictive information when structural morphology is combined with the systemic covariates. Because demographic and socioeconomic covariates are introduced after graph propagation, their information is separated from the intermediate spatial message-passing stages, allowing the model to combine structural and systemic information at the prediction stage.

\subsection{Neuroanatomical Drivers of Cognitive Phenotypes}
Feature attribution using IG identified distinct morphological features that contributed to the model predictions across cognitive targets. The biological interpretation of these features was considered in relation to prior neuroanatomical literature, but the attribution results should be interpreted as model-level associations rather than evidence of causal mechanisms. Driven in part by the global scaling strategy, cortical surface area within the bilaterally massive superior frontal gyrus was among the higher-attribution morphological features across all four tasks. This reflects the region's protracted structural maturation and its central role in broad executive functioning and cognitive control throughout childhood development \cite{mills2016structural}.

Beyond this domain-general frontal involvement, the ST-GNN identified partially distinct, task-specific patterns of structural attribution. Performance on tasks requiring rapid visual attention and cognitive flexibility—specifically the Flanker and Dimensional Change Card Sort assessments—was heavily influenced by cortical thickness in posterior visual regions. The bilateral pericalcarine cortex and left cuneus ranked among the highest anatomical predictors for both tasks. Given that these assessments require subjects to rapidly process visual shift attention between conflicting visual dimensions, the model assigned relatively high attribution to features within primary visual and associative cortical regions. Furthermore, the Flanker task uniquely recruited structural signals from the medial temporal lobe, with bilateral entorhinal and parahippocampal thickness showing relatively high attribution, consistent with the broader involvement of medial temporal regions in contextual processing.

The List Sorting working memory task yielded a markedly distinct anatomical profile. While cortical surface area in the superior frontal gyrus remained relevant, cerebellar cortex volume received the highest attribution among the evaluated structural features. The right and left cerebellar cortices emerged as the most critical localized features for working memory performance. This finding aligns directly with modern connectomic literature, which extends the cerebellum's functional relevance far beyond motor execution, establishing its critical role in verbal working memory, articulatory rehearsal, and sequential information processing \cite{begue2024cerebellum,schmahmann2019dysmetria}.

Finally, the Pattern Comparison processing speed task relied on a morphological network optimized for rapid stimulus evaluation. Alongside the superior frontal gyrus, cortical thickness within the bilateral insula and the surface area of the isthmus of the cingulate gyrus were highly predictive. These regions have previously been implicated in attentional and salience-related processing, providing a possible neurobiological context for their contribution to the model predictions \cite{menon2010saliency}. By identifying these specific morphological variations, the ST-GNN identified partially distinct patterns of regional structural attribution across the four cognitive domains.

\subsection{Methodological Insights into Spatial Message Passing}

An analysis of the Graph Attention distributions revealed systematic differences between attention-based routing scores and predictive feature attribution. Despite possessing minimal predictive importance in the IG attribution maps, central fluid spaces---specifically the inferior lateral ventricles and the fourth ventricle---consistently ranked as primary topological hubs for attention weights across all tasks.

\subsubsection{The Euclidean Bottleneck}
Rather than indicating a biological role for these regions in cognition, this phenomenon highlights a critical architectural artifact of purely spatial GNNs. Because the adjacency matrix is constructed from Euclidean proximity in MNI space, centrally positioned structures can acquire relatively high attention-based routing scores and facilitate information propagation between spatially separated regions. This pattern illustrates how localized spatial message passing can propagate information through intermediate nodes within the fixed graph topology.

\subsubsection{Decoupling Routing from Biological Relevance}
To rigorously evaluate whether the ST-GNN's predictive capacity stems from genuine neurobiological features or these architectural artifacts, we contrasted the initial spatial attention-based routing centrality (hub score from the first GAT block) against predictive feature attribution (Integrated Gradients). As shown in Fig. \ref{fig:gat_vs_ig_scatter}, these metrics exhibit a distinct L-shaped distribution, indicating that attention-based routing scores and predictive feature attribution need not coincide.

The topological hubs (highlighted in red) strictly cluster in the bottom-right quadrant across all four cognitive domains. The observed pattern indicates that these centrally positioned nodes can receive relatively high routing scores while contributing comparatively little to the final prediction according to Integrated Gradients.

Conversely, the high-attribution morphological features (highlighted in blue) anchor the upper-left and upper-center regions of the manifold. Structural features such as the cerebellar cortex volume in working memory and superior frontal morphology in executive-related tasks receive relatively high predictive attribution (high IG attribution) without requiring correspondingly high routing scores (moderate scores in the first GAT block). Ultimately, this confirms that sparse spatial graphs can exploit geographic bottlenecks for highly efficient data propagation without assigning these routing hubs correspondingly high model attribution.

\subsection{Limitations and Future Directions}
This study has several limitations. First, the cross-sectional design does not permit assessment of longitudinal cognitive trajectories or developmental change. Second, the spatial graph is constructed from Euclidean proximity among atlas-defined regional coordinates in standardized MNI space and therefore represents a geometric approximation rather than subject-specific structural or functional connectivity. Third, the use of a fixed atlas and neighborhood definition means that the learned representation may depend on the selected parcellation and graph construction strategy. Although family-aware cross-validation and training-only preprocessing were used to reduce information leakage, the analysis was performed within a single pediatric cohort, and the generalizability of the results to independent cohorts, acquisition protocols, and unseen sites remains to be established. In addition, the target-specific complete-case cohorts may limit generalizability if participants with missing cognitive assessments differ systematically from those retained for analysis. Finally, Integrated Gradients provides model-attribution information but does not establish causal or mechanistic relationships between regional morphology and cognition. Future studies should therefore evaluate external and cross-site validation, alternative spatial representations, longitudinal data, and individualized structural connectivity derived from diffusion MRI.

\section{Conclusion}
\label{sec:conclusion}
This study investigated whether spatially informed representation learning can improve the prediction of individual differences in pediatric cognitive performance from regional structural MRI. We introduced ST-GNN, which represents 101 FreeSurfer-derived anatomical regions as nodes in a shared spatial graph constructed using MNI-space proximity. Regional morphometric features were processed through localized graph message passing and subsequently combined with demographic and socioeconomic covariates through a late-fusion framework.

Across four NIH Toolbox cognitive domains, the results showed that demographic and socioeconomic variables accounted for a substantial proportion of the observed variance, while incorporating structural morphometry provided additional predictive information when the morphology was represented using the proposed spatial graph. In contrast, flattened structural representations and the topology-free MLP provided weaker predictive performance, supporting the value of explicitly incorporating spatial relationships among regional morphometric features. The ablation analyses further indicated that sparse spatial topologies generally performed better than fully connected and randomized alternatives, while comparisons across different neighborhood sizes showed that predictive performance was relatively stable within sparse configurations and varied across cognitive domains. The main ST-GNN configuration used a fixed $K=3$ spatial neighborhood, with alternative values examined as architectural ablations.

The comparison of feature-scaling strategies further showed that global channel-wise standardization, based only on non-zero training measurements pooled across anatomical regions, generally outperformed node-wise standardization in the present setting. From a computational perspective, this finding suggests that retaining absolute between-region differences in morphometric magnitude provides a powerful, representation-level inductive bias for spatial graph learning, offering a highly transferable principle for engineering applications dealing with hierarchical structural data. At the same time, the substantially lower performance of the morphology-only configuration relative to the covariate-fused model indicates that structural morphology and systemic factors provide complementary predictive information for pediatric cognitive performance.

Integrated Gradients identified partially distinct patterns of regional morphological attribution across the four cognitive domains, with relatively high attribution to superior frontal, posterior visual, cerebellar, and insular features, depending on the target. These findings provide model-level evidence that different cognitive phenotypes are associated with distinct patterns of regional structural information within learned representations. The accompanying attention analysis further demonstrated that graph routing scores and predictive feature attribution do not necessarily coincide, highlighting the importance of distinguishing computational message-passing patterns from model-attribution patterns when interpreting spatial graph networks.

Overall, the findings support the use of a fixed, spatially informed graph representation as a practical approach for predicting pediatric cognitive performance from standard T1-weighted structural MRI while incorporating relevant systemic covariates. The framework does not require subject-specific functional or diffusion-derived connectivity and therefore provides a comparatively simple approach for exploring spatially informed structural MRI prediction. However, the present findings are limited to a cross-sectional analysis within the Healthy Brain Network cohort, and the MNI-based spatial graph represents geometric proximity rather than individualized anatomical or functional connectivity. Independent external validation, cross-site evaluation, longitudinal investigation, and further assessment of attribution stability are needed to determine the generalizability and translational relevance of the framework. Thus, the proposed ST-GNN should be viewed as a research framework for spatially informed cognitive prediction rather than as a validated clinical screening or decision-support tool.

\section*{Acknowledgment}
This manuscript was prepared using a limited-access dataset obtained from the Child Mind Institute Biobank, Healthy Brain Network (HBN). This manuscript reflects the views of the authors and does not necessarily reflect the opinions or views of the Child Mind Institute. This work was supported by the Prime Minister's Education Assistance Trust, Government of the People’s Republic of Bangladesh.

\section*{Code \& Data Availability}
Codes associated with this study will be made publicly available after peer review. The structural neuroimaging and phenotypic data utilized in this research were obtained from the pediatric Healthy Brain Network (HBN) cohort. This limited-access dataset is available through the Child Mind Institute Biobank to approved researchers. Access requires an executed data usage agreement, with comprehensive instructions and data access portal links available via the Child Mind Institute (\url{http://fcon_1000.projects.nitrc.org/indi/cmi_healthy_brain_network/}).



\printcredits

\bibliographystyle{elsarticle-num}

\bibliography{cas-refs}



\end{document}